\documentclass[10pt,twocolumn,letterpaper]{article}

\usepackage{cvpr}
\usepackage{times}
\usepackage{epsfig}
\usepackage{graphicx}
\usepackage{amsmath}
\usepackage{amssymb}
\usepackage{booktabs}
 \usepackage{tabularx}
  \usepackage{adjustbox}
 \usepackage{ctable}
 \usepackage{multirow}
 \usepackage{balance}

\usepackage{subcaption}
\newcommand{\ra}[1]{\renewcommand{\arraystretch}{#1}}
\newcommand{\dashrule}[1][black]{%
  \color{#1}\rule[\dimexpr.5ex-.2pt]{4pt}{.4pt}\xleaders\hbox{\rule{4pt}{0pt}\rule[\dimexpr.5ex-.2pt]{4pt}{.4pt}}\hfill\kern0pt%
}

\begin{document}

%%%%%%%%% TITLE
\title{MonoDVPS: A Self-Supervised Monocular Depth Estimation Approach to Depth-aware Video Panoptic Segmentation}

\author{Andra Petrovai\\
Technical University of Cluj-Napoca\\
Cluj-Napoca, Romania\\
{\tt\small andra.petrovai@cs.utcluj.ro}
\and
Sergiu Nedevschi\\
Technical University of Cluj-Napoca\\
Cluj-Napoca, Romania\\
{\tt\small sergiu.nedevschi@cs.utcluj.ro}
}

\maketitle
\thispagestyle{empty}

%%%%%%%%% ABSTRACT
\begin{abstract}
  Depth-aware video panoptic segmentation tackles the inverse projection problem of restoring panoptic 3D point clouds from video sequences, where the 3D points are augmented with semantic classes and temporally consistent instance identifiers. We propose a novel solution with a multi-task network that performs monocular depth estimation and video panoptic segmentation. Since acquiring ground truth labels for both depth and image segmentation has a relatively large cost, we leverage the power of unlabeled video sequences with self-supervised monocular depth estimation and semi-supervised learning from pseudo-labels for video panoptic segmentation. To further improve the depth prediction, we introduce panoptic-guided depth losses and a novel panoptic masking scheme for moving objects to avoid corrupting the training signal. Extensive experiments on the Cityscapes-DVPS and SemKITTI-DVPS datasets demonstrate that our model with the proposed improvements achieves competitive results and fast inference speed. 
\end{abstract}

%%%%%%%%% BODY TEXT
\section{Introduction}

%Environment perception is a fundamental component of autonomous systems such as robotics or automated driving. The goal of perception in the context of automated driving is to detect static road infrastructure, but also to detect, track and classify objects in the driving environment.  

Environment perception is a fundamental component of autonomous systems such as automated vehicles. Traditionally, in order to achieve robust perception, multi-modal sensors such as LiDARs and cameras scan the environment and their output is either fused or processed independently by algorithms in order to detect, track and classify the objects in the environment. While specialized sensors such as LiDARs provide precise depth measurements, they have at the same time a high cost and reduced output density. In a multi-modal sensory setup, further challenges have to be addressed, such as sensor synchronization and fusion. Images can be used to infer both semantics and depth, and as a result, perception using cameras only is attractive due to the simple setup and low cost. 

%Images captured by cameras provide rich appearance information such as color and texture, which are useful cues in inferring the semantics of the environment.

\begin{figure}
	\centering\includegraphics[width=\linewidth]{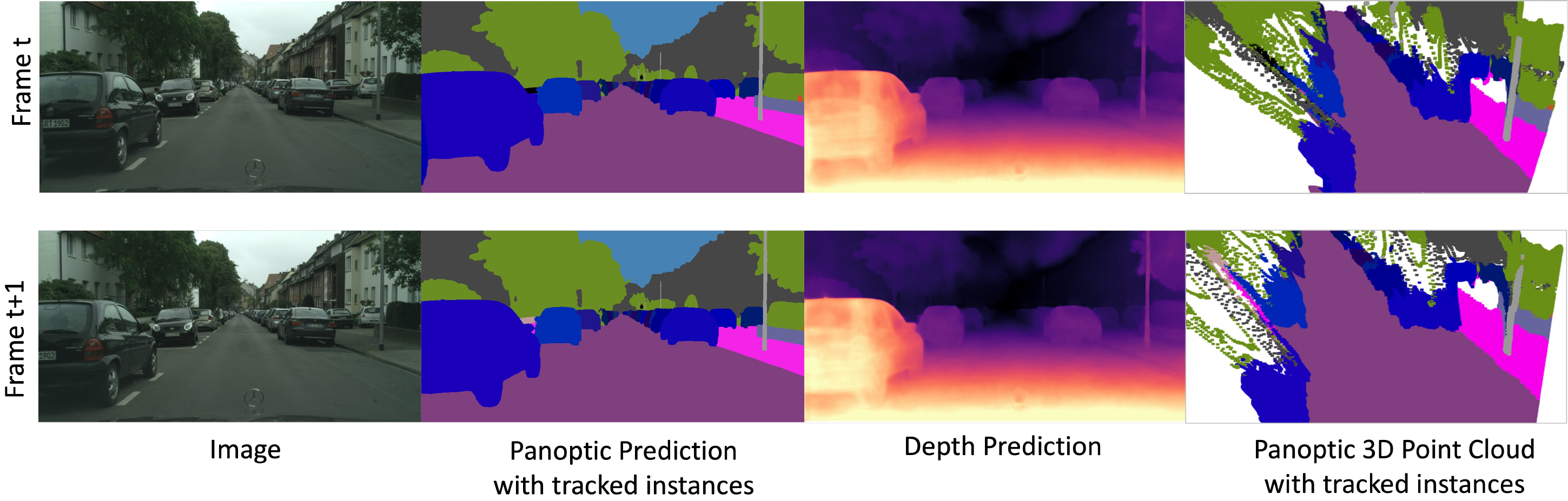}
	\caption{\textbf{Depth-aware Video Panoptic Segmentation.} We generate temporally consistent panoptic 3D point clouds from monocular sequences. Our multi-task network predicts video panoptic segmentation with tracked instances (same instance identifier and color in consecutive frames) and monocular depth. Depth is trained in a self-supervised regime, while video panoptic segmentation is trained in a semi-supervised regime, on both human annotated labels and pseudo-labels. }
\label{fig:dvps}
\end{figure}

Depth estimation from monocular cameras is a long-lasting research field of computer vision. With the advent of deep learning, it has seen major leaps in performance, especially in the supervised setting. However, large-scale acquisition of depth ground truth has a prohibitively large cost, which led to the emergence of self-supervised monocular depth estimation (SSMDE) methods where ground truth is not employed. The idea behind these approaches is that an image synthesis formulation using 3D reprojection models can be used to jointly learn depth and ego motion. SSMDE is usually less accurate than supervised methods, however the gap can be bridged by leveraging large-scale datasets with unlabeled video sequences.

%Depth estimation from stereo or monocular cameras is a long-lasting research field of computer vision. In order to solve stereo reconstruction, the research community has proposed traditional algorithms \cite{scharstein2002taxonomy,hirschmuller2007stereo} and more recently deep convolutional neural networks \cite{zbontar2015computing,kendall2017end}. However, for accurate results the stereo rig has to be carefully calibrated and synchronized. On the other hand, monocular depth estimation is an ill-posed problem since an infinity of 3D scenes can be generated from a single image. With the advent of deep learning, monocular depth estimation from a color image, has seen major leaps in performance, especially in a supervised setting. However, large-scale acquisition of depth ground truth has a prohibitively large cost, which led to the emergence of self-supervised monocular depth estimation (SSMDE) methods, which do not employ ground truth. The idea behind these approaches is that an image synthesis formulation using 3D reprojection models can be used to jointly learn depth and ego motion. 

Panoptic segmentation \cite{panoptic_paper} provides a rich 2D environment representation by performing pixel-level semantic and instance-level segmentation. Video panoptic segmentation \cite{kim2020video} extends the task to video and requires temporally consistent instance predictions. To obtain a holistic 3D representation of the environment, depth-aware video panoptic segmentation (DVPS) \cite{vip_deeplab} is introduced as the combination of monocular depth estimation \cite{saxena2005learning} and video panoptic segmentation. ViP-DeepLab \cite{vip_deeplab} proposes a strong baseline for the task, with a network trained in a supervised regime. 
%The focus is only on  accuracy, therefore the inference time is high.
%The naive approach of employing separate networks for each sub-task is unfeasible from a practical perspective in applications where the hardware resources are limited due the high computational cost and high memory footprint. Multi-task learning, which unifies all the sub-tasks under one architecture, reduces the latency, however it is challenging as it requires careful balancing to ensure high performance for all the sub-tasks. 

With this work, we aim to reduce the ground truth dependency and leverage large amounts of unlabeled video sequences for improved depth-aware video panoptic segmentation (DVPS). Therefore, we propose \textbf{MonoDVPS} a novel multi-task network for the DVPS task based on self-supervised monocular depth estimation and semi-supervised video panoptic segmentation. For video panoptic segmentation, we train on both labeled images and pseudo-labels. In the complex multi-task training setting, we aim to improve the performance of all sub-tasks and propose several techniques for this purpose. We investigate loss balancing to increase the accuracy of all sub-tasks and leverage panoptic guidance to reduce the depth error. Since the self-supervised depth estimation relies on the assumption that the scene is static, moving objects corrupt the training signal and introduce high photometric errors. To overcome this problem, we propose a novel moving objects masking based on panoptic segmentation maps from consecutive frames and remove those pixel locations from the photometric loss computation. To further improve the depth prediction, we introduce three loss terms based on the observation that depth discontinuities occur at panoptic edges: panoptic-guided smoothness loss \cite{saeedan2021boosting} to ensure depth smoothness of neighboring pixels inside panoptic segments, panoptic-guided edge discontinuity loss to enforce large depth difference at panoptic contour and finally we adapt the semantic-guided triplet loss \cite{jung2021fine} into the panoptic domain. We perform extensive experiments on the Cityscapes-DVPS \cite{vip_deeplab} and SemKITTI-DVPS \cite{vip_deeplab} datasets and demonstrate the effectiveness of our approach.

\section{Related Work}

%\subsection{Panoptic Segmentation} Proposal-based methods build on top of an object detection network that generates bounding boxes and overlapping instance masks. The network is usually extended with a semantic segmentation head, and a post-processing step solves overlaps and conflicts between instance and semantic classes. Many works \cite{PanopticFPN,UPSnet,ITSCAndra,AndraITSC2019,SeamlessSegmentation,mohan2021efficientps} employ the two-stage instance segmentation network Mask R-CNN \cite{MaskRCNN} and propose novel merging operations or parameter-free panoptic head. Since these methods are computationally intensive, the research community has proposed more lightweight and fast panoptic segmentation networks \cite{FPSNet,DensePredictions,petrovai2020real,weber2020single,petrovai2022fast} suitable for real-time applications, that employ single-stage object detectors \cite{lin2017focal,FCOS}. The initial performance gap between proposal-based and proposal-free methods has been closed with the introduction of  the proposal-free Panoptic DeepLab \cite{PanopticDeepLab} network, which achieves state-of-the-art results on multiple benchmarks. Axial-DeepLab \cite{wang2020axial} proposes a stand-alone attention model with factorized 2D self-attention layers.
\textbf{Depth-aware Video Panoptic Segmentation.} ViP-DeepLab \cite{vip_deeplab} introduces the task as well as the baseline network. ViP-DeepLab processes concatenated image pairs and extends Panoptic DeepLab \cite{PanopticDeepLab} with a next-frame instance center offset decoder and a monocular depth estimation decoder. The main differences between our network and ViP-DeepLab are related to the depth estimation and tracking sub-tasks. We employ self-supervised depth estimation, which requires no ground truth and is based on geometric projections that allow view-synthesis of adjacent frames. On the other hand, ViP-DeepLab trains the depth in a fully supervised regime with a regression loss. For improved depth results, we adopt multi-scale depth prediction at four scales during training, while ViP-DeepLab uses a single scale. For instance tracking, we use self-supervised optical flow estimation to warp the current panoptic prediction into the next and match the instance IDs based on their mask overlap. On the other hand, ViP-DeepLab tracks instances by predicting pixel-wise offsets to the previous image instance centers, after which it uses the same instance matching algorithm.

\textbf{Video Panoptic Segmentation.} Kim \textit{et al.} has recently introduced the task in \cite{kim2020video} along with the baseline VPSNet network. VPSNet is built on top of the proposal-based two-stage panoptic segmentation network UPSNet \cite{UPSnet}. In order to improve the current prediction, a pixel-level fusion module gathers features from the previous and next five frames, which are further aligned with optical flow and fused with spatio-temporal attention. Our network, on the other hand, does not employ temporal aggregation and operates in an online fashion, it processes only the current frame, which makes the inference faster. For the tracking functionality, VPSNet employs a MaskTrack head \cite{yang2019video}, which learns an affinity matrix between RoI proposals, while our network learns optical flow and uses mask overlapping for instance ID propagation. SiamTrack \cite{woo2021learning} improves VPSNet by designing novel learning objectives that learn segment-wise and dense temporal associations in a contrastive learning framework. In contrast to VPSNet and SiamTrack, our network is box-free and uses the paradigm \textit{segment then group thing pixels into instances}, which makes our network much faster. VPS-Transformer \cite{petrovai2022time} proposes a hybrid architecture derived from Panoptic DeepLab \cite{PanopticDeepLab} with a focus on both efficiency and performance. A video module based on the Transformer block \cite{vaswani2017attention}, equipped with attention mechanisms, models spatio-temporal relations between features from consecutive frames for enhanced feature representations. We do not explicitly encode video correlations between frames, but we employ semi-supervised learning with pseudo-labels for improved video panoptic predictions. 

% For instance tracking, a stitching algorithm is proposed, which temporally propagates instance identifiers across instances with significant overlap.

\textbf{Semantically-guided SSMDE.} The SFMLearner \cite{zhou2017unsupervised} is the first solution for self-supervised depth estimation which jointly trains an ego motion and a depth estimation network. Recent works tackle various problems that arise from the self-supervised problem formulation. For example, Monodepth2 \cite{monodepth2} solves the stationary camera problem by introducing an auto-masking of stationary pixels and the occlusion problem with a minimum reprojection loss. Potentially moving objects break the rigid scene assumption and corrupt the training signal. Semantic and instance information can be used to detect moving objects. Casser \textit{et al.} \cite{casser2019unsupervised} introduces a 3D object motion network that processes images filtered by instance masks. SGDepth \cite{klingner2020self} detects frames with moving objects based on semantic segmentation and removes them from the training set. \cite{tosi2020distilled} segments the object motion with semantic knowledge distillation. Guizilini \textit{et al.} \cite{Guizilini2020Semantically-Guided} enhances the feature representation with semantic guidance from a fixed teacher segmentation network using pixel-adaptive convolutions \cite{su2019pixel}. The tight relation between semantic segments and depth has been exploited in several works by introducing semantic-guided loss functions: semantics-guided smoothness loss \cite{chen2019towards}, which enforces depth smoothness for neighboring pixels in a semantic segment, cross-domain discontinuity loss \cite{zama2018geometry}, which is based on the assumption that pixels across the semantic edges should have large depth differences. A panoptic-guided smoothness loss leveraging ground truth or panoptic predictions is introduced in \cite{saeedan2021boosting}. To improve depth alignment to semantic segmentation, the semantics-guided triplet loss \cite{jung2021fine} with a patch-based sampling technique along semantic edges is proposed. While these networks solely aim to improve the training of a depth estimation network, we train the depth in a multi-task network and we aim to improve all sub-tasks. Compared to semantic edges, panoptic edges, which separate instances but also \textit{stuff} segments, are better aligned to depth edges, where depth discontinuities occur. Therefore, we introduce panoptic-guided loss functions to improve depth training and extend the above semantic-guided losses \cite{chen2019towards, zama2018geometry, jung2021fine} to the panoptic domain. 

\begin{figure*}[t]
	\centering\includegraphics[width=\linewidth]{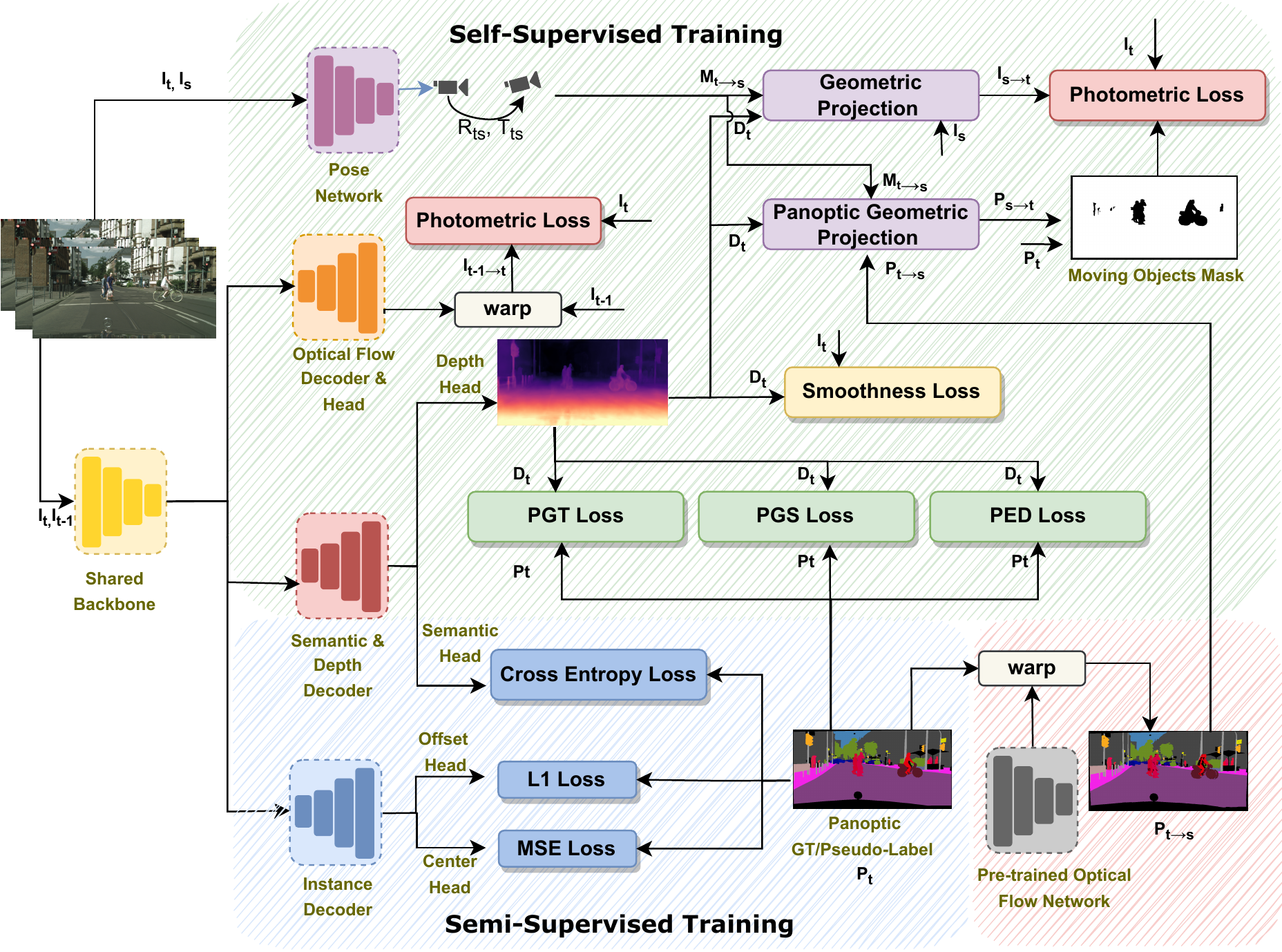}
	\caption{\textbf{Our MonoDVPS Depth-aware Video Panoptic Segmentation Network.} We employ a mixed training regime, where the depth, optical flow and ego motion are trained in a self-supervised manner. Panoptic segmentation is semi-supervised with ground truth and pseudo-labels. We introduce several loss functions, panoptic-guided triplet loss (PGT), panoptic-guided smoothness loss (PGS) and panoptic-guided edge discontinuity loss (PED) to improve the depth training. A novel moving objects mask, computed using the panoptic label, is used to mask the photometric loss.  }
\label{fig:highlevel}
\end{figure*}

\section{Method}

We propose MonoDVPS, a novel depth-aware video panoptic segmentation network that performs panoptic segmentation, instance tracking and monocular depth estimation. In this section we provide details about the network architecture and training framework, illustrated in Figure \ref{fig:highlevel}.

\subsection{Video Panoptic Segmentation}

%VPS-Transformer extends the image panoptic segmentation network Panoptic-DeepLab \cite{PanopticDeepLab} to video with a Transformer video module and an optical flow decoder for instance tracking. 

\textbf{Baseline Network.} We build our solution on top of the panoptic image segmentation network Panoptic DeepLab \cite{PanopticDeepLab} which we extend to video. This network predicts semantic segmentation and groups pixels into instances to obtain the panoptic prediction. Panoptic DeepLab has a shared backbone and dual decoders for semantic and instance segmentation. The instance segmentation branch predicts the instance centers and the offsets from each instance pixel to its corresponding center. In this way, the grouping operation can be easily achieved by assigning \textit{thing} pixels to the closest center based on the predicted offsets. Panoptic DeepLab is supervised by the semantic segmentation loss, instance center regression loss and instance offset regression loss. We extend the network with an optical flow decoder \cite{petrovai2022time} for instance tracking, which is simultaneously trained with the rest of the network in a self-supervised regime by minimizing the photometric loss between the current and warped previous frame. Tracking is achieved by matching the current instance predictions with the warped instance masks from the previous frame using the predicted optical flow.

%Since VPS-Transformer sequentially processes video frames, a video module based on the Transformer block \cite{vaswani2017attention} is introduced between the backbone and decoders in order to aggregate past (memory) and current (query) features for improved feature representation. The Transformer block features a global time-space attention mechanism, as well as a multi-layer perceptron. The Time-Space attention layer extracts temporal and spatial correspondences between every position in the query and the memory. The Transformer video module has a lightweight architecture and improves the panoptic quality with little extra computation. 

\textbf{Semi-supervised Training with Pseudo-Labels.} We employ the Panoptic DeepLab \cite{PanopticDeepLab} image panoptic segmentation network with HRNet-W48 \cite{WangSCJDZLMTWLX19} to generate pseudo-labels for the unlabeled data on the Cityscapes-DVPS train set. The initial train set provides human annotated labels for every 5th frame in a 30 frame video sequence. Following Naive-Student \cite{chen2020naive}, we use test-time augmentations, such as horizontal flips and multi-scale inputs at scales 0.5:2:0.25 in order to improve the pseudo-labels predictions. For Cityscapes-DVPS images,  the ego-car pixels are labeled as void in order to be ignored during training. We also generate dense pseudo-labels for SemKITTI-DVPS images.

\subsection{Self-Supervised Monocular Depth Estimation}

We extend the semantic decoder with a depth prediction head that has a $[5\times5, 64]$ depthwise separable convolution, followed by  bilinear interpolation, concatenation with low-level features and  $[5\times5, 32]$ and $[1\times1, 1]$ convolutions. We adopt multi-scale depth prediction and image reconstruction at four scales with output stride 2, 4, 8 and 16 relative to the original image resolution. In practice, the network learns the disparity, the inverse of depth, as it is more robust \cite{monodepth2}.

The goal of self-supervised monocular depth estimation is to predict the depth map for a single image, while no ground truth is employed in the training phase. The self-supervised depth estimation paradigm is based on geometric projections and is formulated as the minimization of reprojection errors between synthesized adjacent frames and the current frame. During training sequential triplets of frames are required, while during inference a single frame is processed. The mechanism assumes that the scene is static, the camera is moving and all image regions can be reconstructed from neighboring frames. For camera motion, a separate camera pose estimation network is jointly trained. 

Let $I^t$ be target frame for which the depth is predicted, and $I^s$ the adjacent frames, where $s=\{t-1, t+1\}$, captured by a moving camera. The camera pose network estimates the ego motion $M_{t	\rightarrow s}$, that is the 3D translation $ T_{t	\rightarrow s}$ and rotation $R_{t	\rightarrow s}$ between consecutive 3D positions. 

\begin{equation}
M_{t	\rightarrow s} = \begin{bmatrix}  R_{t	\rightarrow s} &  T_{t	\rightarrow s} \\
								0 & 1
\end{bmatrix}
\end{equation}

Let $K$ be the intrinsic matrix that defines the focal length and principal point, which are known for a specific dataset. For a pixel $p$ in the target frame we compute its corresponding 3D point $x$ by backprojection using the predicted target depth map $D_t$. The 3D point is then displaced by the predicted ego motion $M_{t	\rightarrow s}$ to the source 3D position. Its location in the source frame can be obtained by reprojection:

\begin{equation} \label{projection}
p' = \begin{bmatrix}
K|0
\end{bmatrix} 
M_{t	\rightarrow s} \begin{bmatrix}
D_t(p)K^{-1}p \\
1
\end{bmatrix}
\end{equation}

Finally, the target image can be synthesized from the source image by bilinear interpolation \cite{godard2017unsupervised,monodepth2} $I_{s \rightarrow t} = I_s \langle p' \rangle$. During training, the photometric loss between the synthesized images and the target frame is minimized, which is computed as the weighted sum between structural similarity SSIM \cite{wang2004image} and L1 loss:

   \begin{equation}
   pe(I_t, I_{s \rightarrow t}) = \frac{\alpha}{2}(1 - \text{SSIM}(I_t, I_{s \rightarrow t})) + (1 - \alpha)
   \begin{Vmatrix} I_t -I_{s \rightarrow t}
   \end{Vmatrix}_1
\end{equation}

In practice, we adopt the two photometric loss masking schemes from  \cite{monodepth2}. For occlusions we implement the minimum reprojection loss from all source images. To account for the case when the camera is stationary, that can be manifest as 'holes' of infinite depth in the predicted depth map, we filter out pixels where the reprojection error of the synthesized image $I_{s\rightarrow t}$ is lower than the original image $I_s$ \cite{monodepth2}. The photometric loss is computed as the average of the photometric losses at four scales. The predicted lower resolution depth maps at each scale are first upsampled to scale $1/2$ from the original resolution and then are used for reprojection. Based on the assumption that depth discontinuities occur at image edges, we adopt an edge-aware smoothness loss $\mathcal{L}_{sm}$ \cite{monodepth2} that encourages adjacent pixels to have similar depth values unless an image edge is present:

\begin{equation}
\mathcal{L}_{sm} = |
\partial_x \bar{d_t}| e^{-|\partial_x I_t|} +  |
\partial_y \bar{d_t}| e^{-|\partial_y I_t|}
\end{equation}

where $\bar{d_t}$ is the mean normalized inverse depth.

%  \begin{equation}
%  L_p = \min_{s} pe(I_t, I_{s \rightarrow t})
%  \end{equation}

%   \begin{equation}
%  \mu = [\min_{s} pe(I_t, I_{s \rightarrow t}) < \min_{s} pe(I_t, I_s)  ]
%  \end{equation}

%\subsection{Improving Depth with Panoptic Guidance}

%\begin{figure}
%\centering
%\begin{adjustbox}{width=0.7\columnwidth}
%\includegraphics{imgs/panloss.png}
%\end{adjustbox}
%\caption{\textbf{Panoptic-guided losses for improving depth.} Comparison between the baseline ('Depth') and the network trained with panoptic guidance ($\mathcal{L}_{PGS}$, $\mathcal{L}_{PED}$, $\mathcal{L}_{PGT}$).}
%\end{figure}

\subsection{Improving Depth with Panoptic Guidance} 

\begin{figure}[t]
\center
\includegraphics[width=\linewidth]{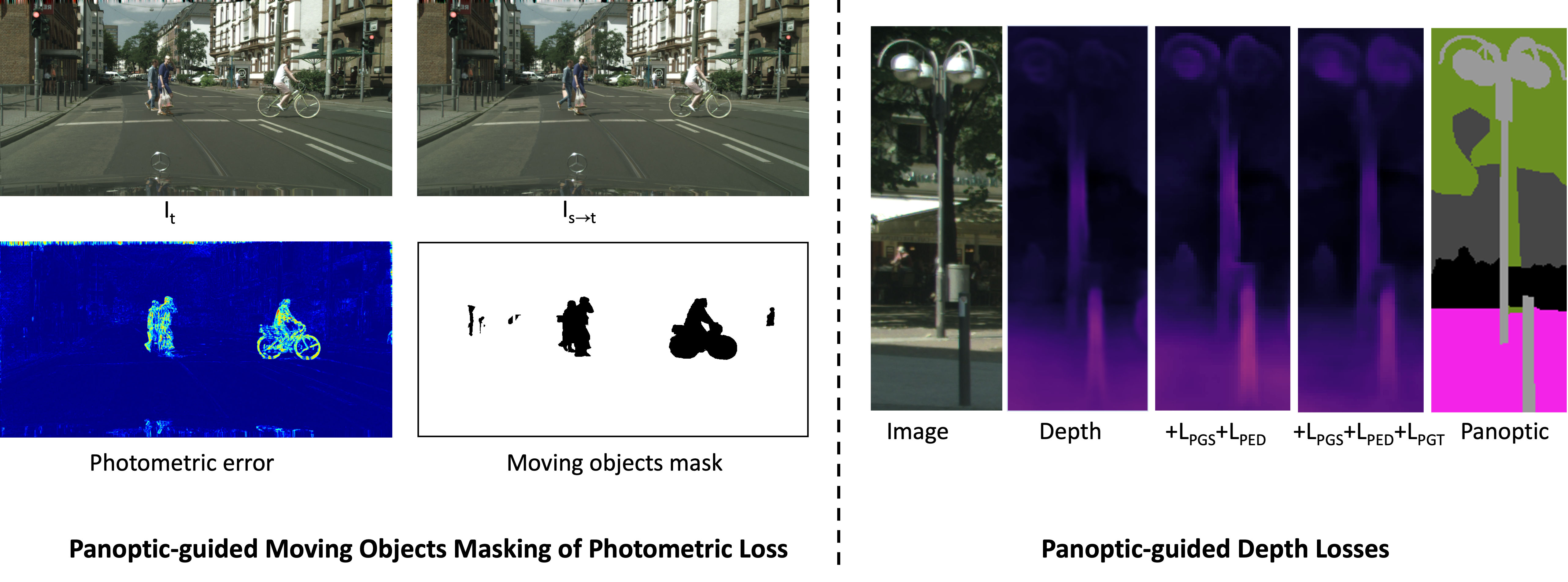}
\caption{\textbf{Moving Objects Masking and Panoptic-Guided Losses.} On the left we illustrate the high photometric loss for moving objects which corrupts the training signal and the moving objects mask. On the right, panoptic-guided depth losses improve the depth prediction. }
\label{pgd}
\end{figure}

We propose two main mechanisms to improve the performance of the depth estimation by panoptic guidance. First, we start from the observation that the panoptic segmentation has a strong correlation with the depth map and introduce three panoptic guided losses. Second, we generate motion masks using consecutive panoptic labels that are applied to the photometric loss. Figure \ref{pgd} presents visual results of the proposed panoptic-guided mechanisms. 

\textbf{Panoptic-guided Smoothness Loss.} We introduce a smoothness loss term \cite{saeedan2021boosting} that enforces similar depth values for adjacent pixels inside a panoptic segment. This loss is derived from $\mathcal{L}_{sm}$, which assumes depth smoothness in the presence of low image gradient. On the other hand, we observe that there is a stronger alignment between depth edges and panoptic contours. To this end, we introduce the following loss:
 
\begin{equation}
\mathcal{L}_{pgs} = |
\partial_x \bar{d_t}| (1 - \partial_x {P_{t}})+  |
\partial_y \bar{d_t}| (1 - \partial_y {P_{t}})
\end{equation}

where $P_{t}$ represents the panoptic ground truth label, $\partial {P_{t}}$ are the panoptic contours and $\bar{d_t}$ is the mean normalized inverse depth. For two adjacent pixels $(p_0, p_1)$, we define the $\partial_x {P_{t}}(p_0, p_1)$ as the Iverson bracket:
%\begin{equation}
%\partial_x {P_{t}}(p_0, p_1) 
% \begin{cases} 
% 1  & P(p_0) \neq  P(p_1)\\
% 0 & P(p_0) = P(p_1)
% 
% \end{cases}       
%\end{equation}

\begin{equation}
\partial_x {P_{t}}(p_0, p_1)  = [P(p_0) \neq  P(p_1)]
\end{equation}

\textbf{Panoptic-guided Edge Discontinuity Loss.} Based on the observation that adjacent pixels across the panoptic edges may have large depth discontinuities, we design the following panoptic-guided edge discontinuity term:

 \begin{equation}
\mathcal{L}_{ped} = 
 \partial_x {P_{t}} e^{-|\partial_x \bar{d_t}|}+  
\partial_y {P_{t}} e^{-|\partial_y \bar{d_t}|} 
\end{equation}

This loss enforces a gradient peak in the disparity map at panoptic edges, when we have different panoptic identifiers for adjacent pixels. It represents an extension of the cross-domain discontinuity loss \cite{zama2018geometry} from the semantic to the panoptic domain and is also similar to the panoptic-guided alignment loss from \cite{saeedan2021boosting}.

\textbf{Panoptic-guided Triplet Loss.} We extend the semantic-guided triplet loss \cite{jung2021fine} to the panoptic domain. The idea behind this loss is that pixels across the panoptic contours should have a large depth difference. The problem with the original formulation that uses semantic contours \cite{jung2021fine} is that instances with the same semantic class belong to one segment and the edges between instances are missing. On the other hand, instance edges are present in the panoptic map and panoptic edges are better aligned to the depth edges. The triplet loss is defined as follows. The panoptic segmentation map is divided into $5 \times 5$ patches and those that do not intersect the panoptic contours are discarded. For the remaining patches a triplet loss is defined. This loss is applied in the feature representation space on the normalized depth feature maps at four scales before the last $[1 \times 1, 1]$ convolution. Features in each patch are grouped in three classes: anchor, positive  $P_i^+$ and negative $P_i^-$. The anchor is located at the center of the patch, while positive features are the ones that have the same panoptic class with the anchor and the negative features are the ones with different panoptic class. The triplet loss increases the L2 distance $d_i^-$ between the anchor and the negative features and reduces the L2 distance  $d_i^+$ to the positive features inside a patch. The triplet loss with a margin $m$ is adopted:

 \begin{equation}
\mathcal{L}_{pgt} = \max(0, d_i^+ + m - d_i^-)
\end{equation}

% \begin{equation}
%d_i^s = \frac{1}{|P_i^s|} \sum_{j \in P_i^s} \sqrt{(F_d(i) - F_d(j))^2},    s \in \{+, -\}
%\end{equation}

% \begin{equation}
%d_i^- = \frac{1}{|P_i^-|} \sum_{j \in P_i^-} \sqrt{(F_d(i) - F_d(j))^2}
%\end{equation}

%where $F_d(i)$ is the depth feature of the anchor, $F_d(j)$ is the depth feature of positive or negative depth features inside a patch.

%The margin $m$ ensures that the distance will not be optimized when the difference between the positive distance and the negative distance exceeds the margin. 

%\begin{tabularx}{\linewidth}{ll}
%\includegraphics[width=.3\linewidth]{imgs/debug_batch_images_0.png}\captionof{figure}{Figure A}\label{fig:taba} &
%\includegraphics[width=.3\linewidth]{imgs/debug_batch_images_0.png} \captionof{figure}{Figure A} \label{fig:taba}
%\end{tabularx}

%\begin{figure}
%\begin{subfigure}{0.48\textwidth}
%\includegraphics[width=\linewidth]{imgs/debug_batch_images_0.png}
%    \caption{Image $I_t$}
%\end{subfigure} \hfill
%\begin{subfigure}{0.48\textwidth}
%\includegraphics[width=\linewidth]{imgs/projected_img.png}   
%    \caption{Image $I_{{s \rightarrow t}}$}\label{fig:tabb}             
%\end{subfigure}\\ %[1em]
%%
%\begin{subfigure}{0.48\textwidth}
%\includegraphics[width=\textwidth]{imgs/rec_loss.png}
%    \caption{Photometric error}\label{fig:tabc}
%\end{subfigure}\hfill
%\begin{subfigure}{0.48\textwidth}
%\includegraphics[width=\textwidth]{imgs/dynamic_mask_left.png} 
%    \caption{Moving objects mask}\label{fig:tabd}               
%\end{subfigure}
%\caption{\textbf{Moving objects masking.} Moving objects may corrupt the training signal for depth estimation with false high photometric loss as seen in (c) due to the fact that objects movement is not modeled in the self-supervised depth estimation paradigm. We address this by masking pixel locations from potentially moving objects (d) in the photometric loss.}
%\label{moving_objects}
%\end{figure}

\textbf{Panoptic-guided Motion Masking.} In self-supervised depth estimation the scene is assumed to be static and only the ego motion is modeled. Because the object motion is not taken into consideration, moving objects corrupt the training signal with false high photometric loss. In order to solve this issue, we propose a novel scheme to detect moving objects based on the panoptic labels of consecutive frames, where instance identifiers are temporally consistent. Our goal is to define a moving object mask, which contains 0 where a potentially moving object is present in the target frame $I_t$ or the geometrically warped source frames $I_{s \rightarrow t}$, and 1 otherwise. In order to compute the moving object mask we employ the panoptic segmentation pseudo ground truth for the target frame. Since our panoptic pseudo-labels are not temporally consistent, we synthesize panoptic labels for the adjacent source frames from the target panoptic label, in order to ensure that an instance has the same identifier across frames. To achieve this, we employ an external pre-trained optical flow network \cite{zhao2020maskflownet} to warp the target panoptic map $P_t$ to source $\hat{P}_{t \rightarrow s}$. The advantage of using optical flow is that it can model both ego and object motion. An occlusion mask $O_{t \rightarrow s} = [\text{exp}(-\lvert I_s - \hat{I}_{t \rightarrow s} \rvert) > r]$ is designed to remove occluded pixels, where $[\cdot]$ is the Iverson bracket. Then we employ the predicted depth and the geometric projection model from equation \ref{projection} to reconstruct the target panoptic map $P_{s \rightarrow t}$ using nearest neighbor interpolation. The reconstructed panoptic map has the following formulation:
\begin{equation}
P_{s \rightarrow t} = (O_{t \rightarrow s} \hat{P}_{t \rightarrow s}) \langle p' \rangle
\end{equation} 

where $p'$ is the location in the source frame of pixel $p$ in the target frame.

Next, we measure the consistency between the reconstructed  $P_{s \rightarrow t}$ and the true $P_t$ target panoptic map filtered by the instance masks which correspond to potentially moving object classes. Since the geometric projection model accounts only for ego-motion, we assume a high level of consistency between $P_{s \rightarrow t}$ and $P_t$ for a static scene and reduced consistency for moving objects. We measure the consistency as the intersection over union (IoU) between instance masks having the same panoptic identifier in $P_{s \rightarrow t}$ and  $P_t$. We define a threshold $T$ for the IoU, such that if the IoU is lower than $T$, then that instance is considered as a moving object. Pixel locations which correspond to moving objects are excluded from the photometric loss computation. In practice, we obtain the best results with a linear scheduling for threshold $T$. Instead of a fixed value, we set an initial threshold $T=0.7$, which linearly decreases with each iteration. The intuition behind this is that, at the beginning we want the network to learn from static pixels, but as the training progresses we allow more noisy samples to account for potential warping errors.

\subsection{Panoptic 3D Point Cloud}

The depth outputs are up to scale and differ from real-world depth values by a scale factor. Also, each depth map requires a different scale factor, as the depth maps are not inter-frame scale consistent. To recover the true depth, common practice \cite{monodepth2} is to perform per-image median scaling: each predicted depth map is scaled with the ratio between the median of the ground truth and the depth prediction. After scaling the depth maps to real-world values, we generate the panoptic 3D point cloud. To obtain the 3D point in the camera coordinate system for each pixel in the image, we backproject the depth map. Since we also have a panoptic output aligned with the depth map, we augment each 3D point with the panoptic identifier of its corresponding pixel, and finally obtain the panoptic 3D point cloud. To completely remove the ground truth dependency, the scale factor can be directly computed from the predicted depth map as the ratio between a known camera height and a computed camera height. The camera height can be computed as the median or average height of all 3D points labeled as road. We adopt the ground truth median scaling for simplicity.

%Self-supervised monocular depth methods suffer from scale ambiguity. To recover the true depth, common practice \cite{monodepth2} is to perform per-image median scaling: each predicted depth map is scaled with the ratio between the median of the ground truth and the prediction. After scaling, we generate the 3D point cloud by backprojection using the intrinsic matrix. Each 3D point is augmented with the panoptic identifier of its corresponding pixel. 

%We follow this practice, but automatic methods for scale computation that do not rely on ground truth \cite{xue2020toward} could be also employed. 

\subsection{Implementation Details}

%We train a depth network with the same architecture with images of half resolution, as common practice \cite{monodepth2}. The depth head is initialized with pretrained weights from the depth network. 

%To obtain the final depth values, the disparity at the highest resolution, which is $1/2$ from the original image, is activated by a sigmoid layer $\sigma$ and converted to depth with $D = 1 / (a \sigma + b)$, where $a$ and $b$ is the scaling interval,  $a = 0.1$ and $b =100$.
%The architecture of the pose estimation network follows \cite{monodepth2}. It has a lightweight ResNet-18 \cite{ResNet} backbone and a decoder that predicts the 6DOF camera pose, the translation vector and rotation matrix, as Euler angles. The input to the camera pose network is the pair of source and target images. The network is supervised only through the photometric loss. During inference we discard the pose estimation network.
%in the range $\pm 0.2$, $\pm 0.2$, $\pm 0.2$ and $\pm 0.1$.

% Augmented images are processed by the network, while for photometric loss computation original images are used.
During training we optimize nine loss functions. The simple approach of adding up the loss terms results is not optimal, so we balance each loss term with a weighting factor in order to control its importance in the final objective:

   \begin{equation}
    \begin{split}
\mathcal{L}_{total} &= \gamma_{depth}\mathcal{L}_{depth} + \gamma_{sem}\mathcal{L}_{sem}  \\
&+ \gamma_{instance}\mathcal{L}_{instance} + \gamma_{optical}\mathcal{L}_{optical}
\end{split}
  \end{equation}
  
  We define the depth loss as a combination of the photometric loss $\mathcal{L}_{photo}$, smoothness loss $\mathcal{L}_{sm}$, panoptic-guided smoothness loss $\mathcal{L}_{pgs}$, panoptic-guided edge discontinuity loss 
$\mathcal{L}_{ped}$ and panoptic-guided triplet loss $\mathcal{L}_{pgt}$.

 \begin{equation}
 \begin{split}
\mathcal{L}_{depth} &= \gamma_{photo}\mathcal{L}_{photo} + \gamma_{sm} \mathcal{L}_{sm} 
+ \gamma_{pgs}\mathcal{L}_{pgs}   \\
&+  \gamma_{ped}\mathcal{L}_{ped} +  \gamma_{pgt}\mathcal{L}_{pgt}
\end{split}
  \end{equation}
  
Following \cite{PanopticDeepLab}, we define $\mathcal{L}_{instance}$ as the weighted sum between mean squared error (MSE) for instance center prediction head and L1 loss for center offset head. Instance weights are the same as in \cite{PanopticDeepLab}. We set $\gamma_{sem} = 1$, $\gamma_{depth} = 50$, $\gamma_{instance} =1$, $\gamma_{optical} = 10$, $\gamma_{photo} =1$, $\gamma_{sm} =0.001$,  $\gamma_{pgs} =0.01$, $\gamma_{ped} = 0.0001$, $\gamma_{pgt} = 0.1$. The weights have been set such that the main losses have a similar magnitude. Details about network training can be found in the supplementary material in Section \ref{impl_sup}.

%    \begin{equation}
%\mathcal{L}_{instance} = \lambda_{center}\mathcal{L}_{center} + \lambda_{offset}\mathcal{L}_{offset}
%  \end{equation}

%  $\lambda_{center} = 200$, $\lambda_{offset} = 0.01$. 

\section{Experiments}

\subsection{Experimental Setup}

\textbf{Datasets.} \textbf{Cityscapes-DVPS} \cite{vip_deeplab} is an urban driving dataset that extends Cityscapes \cite{Cityscapes} to video by providing temporally consistent panoptic annotations to every 5th frame in a 30-frame video snippet. The training, validation, test sets have 2,400, 300 and 300 frames. We extend the training set by generating pseudo-labels for every frame in the video sequence that does not have human annotation. The extended training set contains 14,100 images, 11,700 panoptic pseudo-labels and 2,400 panoptic labels.  \textbf{SemKITTI-DVPS} \cite{vip_deeplab} is based on the odometry split of the KITTI dataset  \cite{geiger2012we,behley2019semantickitti} and provides annotations for every frame in the sequence. The sparse panoptic annotations are obtained by projecting 3D point clouds acquired by LiDAR and augmented with semantic and temporally-consistent instance information to the image plane. The dataset contains 19,020 training, 4,071 validation and 4,342 test images. Annotations are provided for 8 \textit{things} classes and 11 \textit{stuff} classes for both datasets.

\textbf{Evaluation Metrics.} We adopt standard evaluation metrics: Panoptic Quality (PQ) for panoptic segmentation \cite{panoptic_paper}, Video Panoptic Quality (VPQ) for video panoptic segmentation \cite{kim2020video}, Depth-aware Video Panoptic Quality (DVPQ) \cite{vip_deeplab} for both depth and video panoptic. For depth estimation we evaluate using absolute relative error (absRel), squared relative error (sqRel) and root mean squared error (RMS) \cite{eigen2014depth}. We measure the inference time of the network with all the post-processing steps, with batch size of one on a NVIDIA Tesla V100 GPU.

\subsection{Cityscapes-DVPS Results}
In this section, we provide ablation studies and comparison to state-of-the-art on the Cityscapes-DVPS dataset.

\begin{figure*}[t]
\centering
\includegraphics[width=\textwidth]{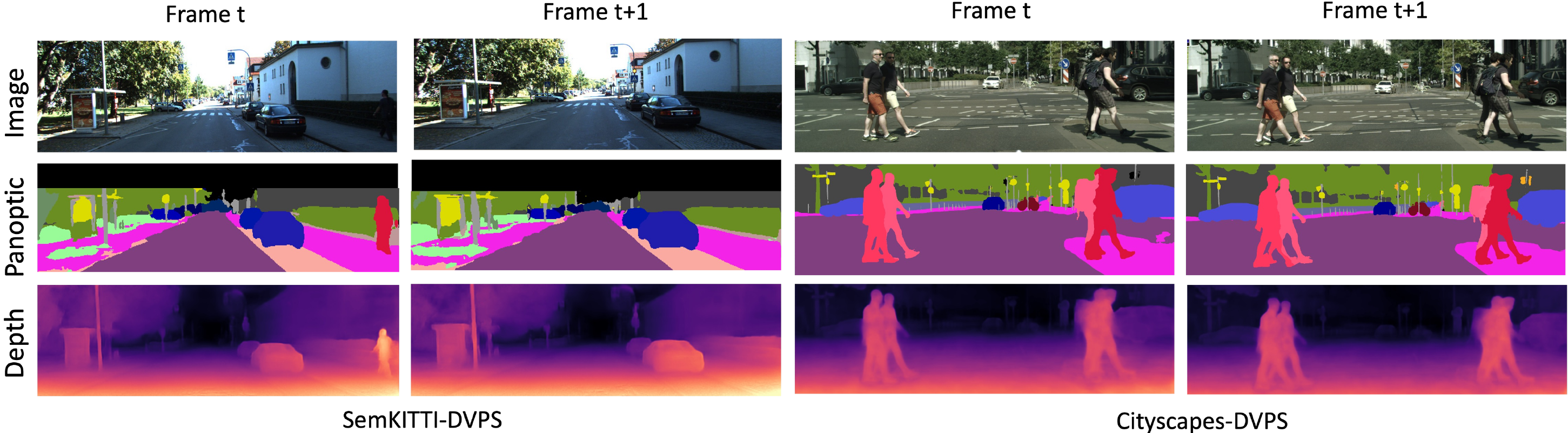}
\caption{\textbf{Qualitative Results.} Video panoptic and depth predictions on SemKITTI-DVPS and Cityscapes-DVPS. }
\label{vis}
\end{figure*}

\begin{table}[t]
\center	
\ra{0.9}

\begin{adjustbox}{max width=\columnwidth}
\begin{tabular}{@{}l|c|c}
%\specialrule{.1em}{.05em}{.05em} 
\toprule
Model                 & PQ $\uparrow $ & absRel  $\downarrow $  \\ \midrule
Panoptic only  & 63.5 & -                              \\
Depth only & - & 0.098                              \\ \midrule
MTL Baseline $\gamma_{depth}=1$ & 62.9 & 0.151 \\ 
+ Loss Balancing  $\gamma_{depth}=50$   & 63.6 & 0.102 \\ 
+ Loss Balancing  $\gamma_{depth}=100$   & 63.2 & 0.102 \\  \midrule
+ Panoptic-guided depth & 63.6 & 0.098 \\  \midrule
\textbf{+ Extended train set} & \textbf{66.5} & \textbf{0.082} \\
\bottomrule
\end{tabular}
\end{adjustbox}
\caption{\textbf{Multi-task Learning (MTL).} Comparison between single task and several multi-task training settings. }
\label{mtl}
\end{table}

\textbf{Multi-task Learning Ablation.} In Table \ref{mtl}, we report the results of single-task baselines and our multi-task network. In the multi-task learning baseline setting, with depth loss weight $\gamma_{depth}=1$, we report a loss in accuracy for both panoptic segmentation and depth compared to single-task baselines. We set $\gamma_{depth} = 50$, as this yields the best PQ and absRel metrics. To further improve the depth prediction, we adopt panoptic-guided losses during training and reach the single-task depth performance. Self-supervised depth estimation does not use depth ground truth during training, therefore can be trained on large-scale unlabeled datasets. By training on the extended train set with panoptic pseudo-labels, we obtain major improvements for both tasks. We present panoptic and depth visualizations on two consecutive frames in Figure \ref{vis}.

% First, we train the panoptic segmentation network Panoptic-DeepLab \cite{PanopticDeepLab} on the Cityscapes-DVPS \texttt{train} set and obtain a score of $63.5\%$ PQ. Our network with the segmentation decoder and depth head is trained for depth estimation in a self-supervised regime on the \texttt{train} set and achieves 0.098 absRel. In the multi-task learning baseline setting, with depth loss weight $\lambda_{depth}=1$, we report a loss in accuracy for both panoptic segmentation and depth. We observe an increase of depth error due to the fact that the network has been initialized with weights from \cite{PanopticDeepLab} trained on Cityscapes for panoptic segmentation. 
% Loss balancing, which sets the depth weight to 50, reduces the depth error to 0.102 absRel. To further improve the depth prediction, we adopt panoptic-guided losses during training and reach the single-task depth performance. Self-supervised depth estimation does not use depth ground truth during training, therefore can be trained on large-scale unlabeled datasets. By training on the extended train set (\texttt{train-sequence}) with panoptic pseudo-labels, we obtain major improvements for both tasks, PQ increases with 3.1$\%$ and the absRel error is reduced from 0.098 to 0.082. 

  \begin{table}[t]
\center	
\ra{0.9}

\begin{adjustbox}{max width=\textwidth}
\begin{tabular}{@{}l|c|c|c}
%\specialrule{.1em}{.05em}{.05em} 
\toprule
Model    & absRel $\downarrow $ & sqRel $\downarrow $ & RMS $\downarrow $ \\  \toprule %\specialrule{.1em}{.05em}{.05em} 
Self-Supervised Depth only & 0.098 & 0.731 & 4.919 \\ \toprule 
MTL Supervised Depth & 0.070 & 0.368 & 3.675 \\ \toprule
MTL Self-Supervised Depth & 0.106 & 0.841  & 5.270  \\  
+ Loss Balancing & 0.102 & 0.767 & 5.034 \\
%+ $\mathcal{L}_{PGS}$ & 0.101 & 0.781 & 5.010 \\
%+ $\mathcal{L}_{PED}$ &  0.101 & 0.778 & 5.024 \\
%+  $\mathcal{L}_{PGT}$ & 0.100 & 0.757 & 4.998 \\ \midrule
%+ $\mathcal{L}_{PGS}$ + $\mathcal{L}_{PED}$ & 0.101 &  0.789  &   4.997 \\
+  $\mathcal{L}_{PGS}$ + $\mathcal{L}_{PED}$ + $\mathcal{L}_{PGT}$ & 0.099 & 0.747  &   4.988 \\ 
+ Moving Objects Masking & 0.098 & 0.701 & 4.864 \\ 
\textbf{+ Extended dataset} & \textbf{0.082} &  \textbf{0.515}  &   \textbf{4.198} \\
\bottomrule
\end{tabular}
\end{adjustbox}
\caption{\textbf{Panoptic-guided Depth Evaluation in a Multi-task Setting.} Ablation study for loss balancing, panoptic-guided depth losses and moving objects masking. }
\label{pan_depth}
\end{table}

%\begin{table}[h!]
%\center	
%\ra{0.9}
%\begin{adjustbox}{max width=\columnwidth}
%\begin{tabular}{@{}l|c|c}
%%\specialrule{.1em}{.05em}{.05em} 
%\toprule
%IoU threshold            & PQ $\uparrow $ & absRel $\downarrow $  \\  \midrule
%0.3 & 63.2 & 0.099 \\
%0.5 & 63.5 & 0.098 \\
%0.7 & 63.9 & 0.102 \\
%\textbf{linear} & \textbf{63.6} & \textbf{0.098}  \\
%\bottomrule
%\end{tabular}
%\end{adjustbox}
%\caption{\textbf{Moving Object Masking.} Ablation study on the IoU threshold used to determine if an object is moving. \textit{Linear} means that the IoU is decreased linearly from 0.7 with each training iteration.}
%\label{mom}
%\end{table}

\textbf{Panoptic-guided Depth Ablation.} In Table \ref{pan_depth} we evaluate the depth under different settings. First, we compare the depth estimation trained in a supervised vs self-supervised regime on the train set in a multi-task setting. For supervised training we formulate depth estimation as regression and adopt the scale-invariant log loss from \cite{eigen}. As expected, supervised depth outperforms self-supervised depth in all metrics. To bridge the gap, we propose several improvements in the self-supervised training process. We balance the multi-task loss by increasing the depth weight and introduce panoptic-guided losses $\mathcal{L}_{PGS}$, $\mathcal{L}_{PED}, \mathcal{L}_{PGT}$ to reduce the depth error. In order to avoid corrupting the training signal in the moving objects region, we design a moving objects masking scheme that further increases the performance. We finally extend the training set (2,400 to 14,410 frames), which significantly reduces the error, showing that a large dataset is a very important element in self-supervised depth training. 
The supplementary material contains ablation studies for the panoptic-guided losses in Table \ref{pan_depth2_sup} and for the moving objects masking in Table \ref{mom_sup}.

\begin{table*}[t]
\ra{0.9}
\begin{adjustbox}{max width=\textwidth}
\begin{tabular}{@{}lccccc}
\toprule 
%\multicolumn{1}{l|}{\multirow{2}{*}{\textbf{Models}}} &   %&\multicolumn{4}{c|}          % & %\multicolumn{4}{c|}{$ \text{VPQ}_k /\text{VPQ}_{k}^{Th} / \text{VPQ}_{k}%^{St}$ for temporal window size $k$}                                                                                                                              
%\multicolumn{1}{l|}{}                                              & \multicolumn{1}{c|}{k = 1}              & \multicolumn{1}{c|}{k = 5}               & \multicolumn{1}{c|}{k = 10}             & \multicolumn{1}{c|}{k = 15}             & \multicolumn{1}{c|}{Average}                            &       Time (ms)                     \\ \hline

\multicolumn{1}{l|}{DVPQ$_{\lambda}^k$ on Cityscapes-DVPS}                            & \multicolumn{1}{c|}{k = 1} & \multicolumn{1}{c|}{k = 2}  & \multicolumn{1}{c|}{k = 3} & \multicolumn{1}{c|}{k = 4} & \multicolumn{1}{c}{DVPQ Average}                              \\ \midrule

\multicolumn{1}{l|}{MonoDVPS $\lambda = 0.50$}                            & \multicolumn{1}{c|}{ 65.9 $|$ 55.7 $|$ 73.3 } & \multicolumn{1}{c|}{59.0 $|$ 43.0 $|$ 70.6}  & \multicolumn{1}{c|}{55.8 $|$  36.9 $|$  69.5} & \multicolumn{1}{c|}{53.5 $|$ 32.5 $|$ 68.8} & \multicolumn{1}{c}{58.6 $|$ 42.0 $|$ 70.6} \\

\multicolumn{1}{l|}{MonoDVPS $\lambda = 0.25$}                            & \multicolumn{1}{c|}{59.3 $|$ 45.4 $|$ 69.4} & \multicolumn{1}{c|}{53.0 $|$ 34.2 $|$ 66.7}  & \multicolumn{1}{c|}{50.2 $|$ 28.9 $|$ 65.7} & \multicolumn{1}{c|}{48.5 $|$ 26.1 $|$ 64.7} & \multicolumn{1}{c}{52.8 $|$ 33.7 $|$ 66.7}   \\

\multicolumn{1}{l|}{MonoDVPS $\lambda = 0.10$}                            & \multicolumn{1}{c|}{39.0 $|$ 23.7 $|$ 50.0} & \multicolumn{1}{c|}{35.1 $|$ 17.5 $|$ 47.9}  & \multicolumn{1}{c|}{33.4 $|$ 14.5 $|$ 47.1} & \multicolumn{1}{c|}{32.5 $|$ 13.1 $|$ 46.6} & \multicolumn{1}{c}{35.0 $|$ 17.2 $|$ 48.0}   \\

\multicolumn{1}{l|}{ MonoDVPS Average } & \multicolumn{1}{c|}{54.7 $|$ 41.6 $|$ 64.2} & \multicolumn{1}{c|}{49.0 $|$ 31.6 $|$ 61.7}  & \multicolumn{1}{c|}{46.5 $|$  26.8 $|$ 60.8} & \multicolumn{1}{c|}{44.8 $|$ 23.9 $|$ 60.0} & \multicolumn{1}{c}{\textbf{48.8 $|$ 31.0 $|$ 61.7}} \\ \midrule  %& 86            
\multicolumn{1}{l|}{MonoDVPS S-MDE} & \multicolumn{1}{c|}{57.2 $|$  48.4 $|$ 63.6} & \multicolumn{1}{c|}{51.0 $|$ 37.0 $|$ 61.0}  & \multicolumn{1}{c|}{47.9 $|$ 31.0 $|$ 60.0} & \multicolumn{1}{c|}{45.7 $|$ 27.0 $|$ 59.3} & \multicolumn{1}{c}{\textbf{50.4 $|$ 35.9 $|$ 61.0}}  \\ \midrule 
 \multicolumn{1}{l|}{ViP-DeepLab (WR-41) \cite{vip_deeplab}} & \multicolumn{1}{c|}{61.9 $|$ 55.9 $|$ 66.3} & \multicolumn{1}{c|}{55.6 $|$ 44.3 $|$ 63.8}  & \multicolumn{1}{c|}{52.4 $|$ 38.4 $|$ 62.6} & \multicolumn{1}{c|}{50.4 $|$ 34.6 $|$ 61.9} & \multicolumn{1}{c}{\textbf{55.1 $|$ 43.3 $|$ 63.6}} \\ %\bottomrule \\
\multicolumn{1}{l|}{ViP-DeepLab* (ResNet-50) \cite{ViP-DeepLab}} & \multicolumn{1}{c|}{47.4 $|$ 38.8 $|$ 53.7} & \multicolumn{1}{c|}{44.0 $|$ 28.1  $|$ 51.6}  & \multicolumn{1}{c|}{39.0 $|$ 23.3 $|$ 50.5} & \multicolumn{1}{c|}{37.5 $|$ 20.2 $|$ 50.0} & \multicolumn{1}{c}{\textbf{42.0 $|$ 27.6 $|$ 51.5}} \\ \bottomrule \\  %& 86     \\
 \end{tabular}
\end{adjustbox}

\begin{adjustbox}{max width=\textwidth}
\begin{tabular}{@{}llccccc}
\toprule
 \multicolumn{1}{l|}{DVPQ$_{\lambda}^k$ on SemKITTI-DVPS}                            & \multicolumn{1}{c|}{k = 1} & \multicolumn{1}{c|}{k = 5}  & \multicolumn{1}{c|}{k = 10} & \multicolumn{1}{c|}{k = 20} & \multicolumn{1}{c}{DVPQ Average}                              \\ \midrule

\multicolumn{1}{l|}{MonoDVPS $\lambda = 0.50$}                            & \multicolumn{1}{c|}{48.7 $|$ 44.7 $|$ 51.7 } & \multicolumn{1}{c|}{ 43.0 $|$ 33.3 $|$ 50.0}  & \multicolumn{1}{c|}{41.6 $|$ 30.7 $|$ 49.6} & \multicolumn{1}{c|}{40.4  $|$  28.4  $|$  49.2 } & \multicolumn{1}{c}{ 43.4 $|$ 34.2 $|$ 50.1} \\

\multicolumn{1}{l|}{MonoDVPS $\lambda = 0.25$}                            & \multicolumn{1}{c|}{45.3 $|$ 39.7 $|$ 49.4} & \multicolumn{1}{c|}{39.8 $|$ 28.9 $|$ 47.8}  & \multicolumn{1}{c|}{38.5 $|$ 26.7 $|$ 47.2} & \multicolumn{1}{c|}{37.6 $|$ 25.0 $|$ 46.8} & \multicolumn{1}{c}{40.3 $|$ 30.0 $|$ 47.8}   \\

\multicolumn{1}{l|}{MonoDVPS $\lambda = 0.10$}                            & \multicolumn{1}{c|}{35.9 $|$ 28.0 $|$ 41.6} & \multicolumn{1}{c|}{ 31.6 $|$ 20.0  $|$ 40.0}  & \multicolumn{1}{c|}{30.6 $|$ 18.4 $|$ 39.4} & \multicolumn{1}{c|}{29.8 $|$ 17.3 $|$ 39.0} & \multicolumn{1}{c}{32.0 $|$ 21.0 $|$ 40.0}   \\

\multicolumn{1}{l|}{ MonoDVPS Average } & \multicolumn{1}{c|}{43.3 $|$ 37.5 $|$ 47.5} & \multicolumn{1}{c|}{ 38.1 $|$ 27.4 $|$ 46.0}  & \multicolumn{1}{c|}{36.9 $|$ 25.2  $|$ 45.4 } & \multicolumn{1}{c|}{ 36.0 $|$  23.6 $|$ 45.0 } & \multicolumn{1}{c}{\textbf{38.6 $|$ 28.4 $|$ 46.0 }} \\ \midrule  %& 86                        
 \multicolumn{1}{l|}{ViP-DeepLab \cite{vip_deeplab}} & \multicolumn{1}{c|}{48.9 $|$ 42.0 $|$ 53.9} & \multicolumn{1}{c|}{45.8 $|$ 36.9 $|$ 52.3}  & \multicolumn{1}{c|}{44.4 $|$ 34.6 $|$ 51.6} & \multicolumn{1}{c|}{43.4 $|$ 33.0 $|$ 51.1} & \multicolumn{1}{c}{\textbf{45.6 $|$ 36.6 $|$ 52.2}} \\ \bottomrule  %& 86     \\
 \end{tabular}
\end{adjustbox}

\caption{\textbf{Depth-aware Video Panoptic Segmentation on Cityscapes-DVPS and SemKITTI-DVPS.} Each cell shows DVPQ$_{\lambda}^k |$ DVPQ$_{\lambda}^k$-Things $|$ DVPQ$_{\lambda}^k$-Stuff. $k$ is the number of frames and $\lambda$ is the threshold of relative depth error. MonoDVPS S-MDE is our network trained in a fully supervised regime for both panoptic and depth. Our networks use the ResNet-50 backbone. ViP-DeepLab uses the heavier WR-41 backbone, Mapillary Vistas pretraining and test-time augmentations. ViP-DeepLab* with the ResNet-50 backbone is evaluated using the author's code and pretrained model.}
\label{dvps}
\end{table*}

\textbf{Depth-aware Video Panoptic Segmentation.} We evaluate DVPQ on Cityscapes-DVPS in Table \ref{dvps}. As expected, DVPQ decreases for larger window size $k$, as the temporal consistency is reduced, and for lower threshold $\lambda$ on depth absRel. We observe a larger performance drop in DVPQ-Things than DVPQ-Stuff when decreasing $\lambda$ for all $k$, which suggests that depth errors are larger on instances than on \textit{stuff} pixels. We also train MonoDVPS in a fully supervised regime for depth and video panoptic segmentation (MonoDVPS S-MDE) and obtain higher DVPQ than the multi-task network trained in a self-supervised depth regime (MonoDVPS Average). Compared to MonoDVPS S-MDE,  depth errors are higher on instances and lower on \textit{stuff} pixels, as indicated by the lower DVPQ-Things and higher DVPQ-Stuff. Our network surpasses ViP-DeepLab with ResNet-50 \cite{ViP-DeepLab} and sets a new state-of-the-art. We present more DVPS results in the supplementary material in Table \ref{dvpq_cs_sup}.

\textbf{Video Panoptic Segmentation.} In Table \ref{vpq}, we compare our MonoDVPS results with the state-of-the-art for video panoptic segmentation. ViP-DeepLab \cite{vip_deeplab} with WR-41 \cite{chen2020naive} backbone and Mapillary Vistas  \cite{Mappilary} pretraining has been designed for accuracy and achieves state-of-the-art results, however it is slow due to the costly test-time augmentations. When using the same ResNet-50 backbone, our network surpasses all other networks, including ViP-DeepLab.

%\begin{table}[t]
%\center	
%\ra{0.9}
%\begin{adjustbox}{max width=\columnwidth}
%\begin{tabular}{@{}l|c|c|c|c|c}
%%\specialrule{.1em}{.05em}{.05em} 
%\toprule
%Model  & Backbone & PQ $\uparrow $ & VPQ $\uparrow $ & DVPQ   $\uparrow $ & Time (s) $\downarrow $  \\ \midrule
%%\specialrule{.1em}{.05em}{.05em}
%VPSNet \cite{kim2020video} & ResNet-50 & 62.7 & 56.1 & - & 0.77 \\ 
%Siam-Track \cite{woo2021learning} & ResNet-50 & 64.6 & 57.3 & - & 0.22 \\
%VPS-Transformer \cite{petrovai2022time} & ResNet-50 & 63.8 & 57.3 & - & \textbf{0.11} \\ 
%ViP-DeepLab \cite{vip_deeplab} & WR-41 & \textbf{70.4} & \textbf{63.1} & \textbf{55.1} & $~$ 54* \\ \midrule
%MonoDVPS (S-MDE) & ResNet-50 & 63.9 & 55.4 & 47.1 & \textbf{0.11} \\
%MonoDVPS (SS-MDE) - baseline & ResNet-50 & 63.6 & 54.9 & 43.5 & \textbf{0.11}\\
%MonoDVPS (ours) + improvements & ResNet-50 & 66.5 & 59.1 & 48.8 & \textbf{0.11}\\
%\bottomrule
%\end{tabular}
%\end{adjustbox}
%\caption{\textbf{Comparison to the state-of-the-art on Cityscapes-DVPS.}  We include both our network with supervised depth (S-MDE) and self-supervised depth (SS-MDE). For ViP-DeepLab* we measure the inference time with the author's code.}
%\label{comparison}
%\end{table}

\subsection{SemKITTI-DVPS Results} 

We evaluate our MonoDVPS network on the SemKITTI-DVPS dataset in Table \ref{dvps}. ViP-DeepLab with WR-41 backbone and test-time augmentations surpasses our results, however our network would also benefit from heavier backbone and these costly operations. Compared to our results on Cityscapes-DVPS, we observe that as the absolute relative depth threshold $\lambda$ decreases, DVPQ$_{k}^{\lambda}$ drops are smaller. For example, on Cityscapes-DVPS the difference DVPQ$_{1}^{0.5}$ - DVPQ$_{1}^{0.25}$ is 6.6\%, while on SemKITTI-DVPS it is 3.4\%. The effect is even more pronounced for smaller $\lambda$. A reason why Cityscapes-DVPS is more sensitive to $\lambda$ could be that the dataset is more complex, with a larger number of instances per image and more difficult scenarios, and our depth has higher errors on instances than on background. 

\begin{table}
\center	
\ra{0.9}
\begin{adjustbox}{max width=\columnwidth}
\begin{tabular}{@{}l|c|c|c|c|c|c|c}
%\specialrule{.1em}{.05em}{.05em} 
\toprule
Model & Backbone & k = 1 & k = 2  & k = 3 & k = 4  & VPQ  $\uparrow $ & Time (s)  \\ \midrule
VPSNet \cite{kim2020video} & ResNet-50 & 62.7 & 56.9 & 53.3 & 51.3 & 56.1 & 0.77  \\ 
Siam-Track \cite{woo2021learning} & ResNet-50 & 64.6 & 57.6 & 54.2 & 52.7 & 57.3 & 0.22  \\	
VPS-Transformer \cite{petrovai2022time}& ResNet-50 & 64.8 & 57.6 & 54.4 & 52.2 & 57.3 & 0.11 \\
ViP-DeepLab* \cite{ViP-DeepLab} & ResNet-50 & 60.6 & 53.1 & 49.9 & 47.7 & 52.8 & -   \\
ViP-DeepLab \cite{vip_deeplab}& WR-41 & \textbf{70.4} & \textbf{63.6} & \textbf{60.1} & \textbf{58.1} & \textbf{63.1} & ~54* \\ \midrule
MonoDVPS (ours) & ResNet-50 & 66.5 & 59.6  & 56.3 & 54.0 & 59.1 & 0.10 \\ \bottomrule
\end{tabular}
\end{adjustbox}
\caption{\textbf{Video Panoptic Segmentation.} $k$ is the window size used for evaluation.  In this paper $k = \{1, 2, 3, 4\}$ is equivalent to  $k = \{1, 5, 10, 15\}$ from \cite{kim2020video,woo2021learning,petrovai2022time}. ViP-DeepLab* is evaluated with the author's code.}
\label{vpq}
\end{table}

\section{Conclusions} 

In this work, we have developed a novel multi-task network for depth-aware video panoptic segmentation with mixed training regimes: self-supervised depth estimation and semi-supervised video panoptic segmentation. By leveraging large amounts of unlabeled images, we improve the performance of both tasks. To further reduce the depth error, we introduce panoptic guidance during training with panoptic-guided losses and a novel panoptic motion masking. The final model achieves competitive performance to the state-of-the-art and offers a good trade-off between inference speed and accuracy.

\section*{Acknowledgement}
This work was supported by the "DeepPerception - Deep Learning Based 3D Perception for Autonomous Driving" grant funded by Romanian Ministry of Education and Research, code PN-III-P4-PCE-2021-1134.

\balance
{\small
\bibliographystyle{ieee_fullname}
\bibliography{egbib}
}

\newpage

\section*{Supplementary Material}

\subsection*{Depth-aware Video Panoptic Segmentation} \label{dvps_sup}

In Table \ref{dvpq_cs_sup} we present a comparison between our MonoDVPS network and concurrent work ViP-DeepLab \cite{ViP-DeepLab} on the DVPS task. We train our network on the original Cityscapes-DVPS training set and obtain 43.4 DVPQ, while on the extended dataset with panoptic pseudo-labels, we achieve 48.8 DVPQ. We surpass ViP-DeepLab \cite{ViP-DeepLab} with the ResNet-50 backbone on the DVPQ score, while having a fast inference speed.

\begin{table}[h!]
\center	
\ra{0.9}
\begin{adjustbox}{max width=\columnwidth}
\begin{tabular}{@{}l|c|c|c|c|c}
%\specialrule{.1em}{.05em}{.05em} 
\toprule
Model & Backbone & DVPQ & DVPQ-Things & DVPQ-Stuff & Time (s)  \\ \midrule
MonoDVPS & ResNet-50 & 48.8 & 31.0 & 61.7  & 0.11 \\
MonoDVPS* & ResNet-50 & 43.4 & 26.2 & 55.9  & 0.11 \\
ViP-DeepLab \cite{ViP-DeepLab} & ResNet-50 & 42.0 & 27.6 & 51.5  & - \\
%ViP-DeepLab \cite{vip_deeplab} & WR-41 & 55.1 & 43.3 & 63.6  & 54 \\
 \bottomrule
\end{tabular}
\end{adjustbox} \\
\caption{\textbf{DVPS on Cityscapes-DVPS.} MonoDVPS* is our network trained on the reduced training set (without extension). 
ViP-DeepLab with ResNet-50 was evaluated with the author's code \cite{ViP-DeepLab}. }
\label{dvpq_cs_sup}
\end{table}

\subsection*{Panoptic-guided Moving Object Masking for Improved Depth Ablation}  \label{moving_objects_sup}
For each instance in frame $t$, we measure the IoU between its mask in the reconstructed panoptic label $P_{s \rightarrow t}$ and its mask in $P_t$. The geometric projection model used for generating  $P_{s \rightarrow t}$ assumes the scene is static and considers only the ego-motion. Therefore, we observe a high overlap between instance masks  for static objects and low overlap for moving objects, since object motion was not modeled. We set a threshold $T$ such that if the IoU is lower than the threshold, the instance is considered a moving object and the pixels corresponding to its mask will be ignored in the photometric loss computation. In Table \ref{mom_sup}, we experiment with $T = \{0.3, 0.5, 0.7\}$ and a linear scheduling. We need to consider that errors from warping with optical flow, geometric reconstruction or occlusions could influence the IoU computation. In consequence, a high threshold $T=0.7$ removes too many instances, while a low threshold $T=0.3$ is too permissive. The linear scheduling obtains the best balance between panoptic and depth performance, with $T=0.5$ being a close second. 

\subsection*{Panoptic-guided Depth Losses Ablation} \label{pgl_ablation_sup}

In Table \ref{pan_depth2_sup} we perform an extensive ablation study for depth estimation. We evaluate the depth output of our multi-task depth-aware panoptic segmentation network on the Cityscapes-DVPS dataset. Specifically, we introduce three panoptic-guided depth losses and evaluate their individual contributions. As seen in Table \ref{pan_depth2_sup}, the panoptic-guided triplet loss $\mathcal{L}_{PGT}$ brings the largest improvement compared to the other two panoptic-guided losses $\mathcal{L}_{PGS}$ and $\mathcal{L}_{PED}$. This could be because $\mathcal{L}_{PGT}$ is less sensitive to errors in the panoptic predictions due to its patch-based formulation. However, we obtain the best results when all three losses $\mathcal{L}_{PGS}$, $\mathcal{L}_{PED}$, $\mathcal{L}_{PGT}$ are used during training.

\begin{table}[h!]
\center	
\ra{0.9}
\begin{adjustbox}{max width=\columnwidth}
\begin{tabular}{@{}l|c|c}
%\specialrule{.1em}{.05em}{.05em} 
\toprule
IoU threshold            & PQ $\uparrow $ & absRel $\downarrow $  \\  \midrule
0.3 & 63.2 & 0.099 \\
0.5 & 63.5 & 0.098 \\
0.7 & 63.9 & 0.102 \\
\textbf{linear} & \textbf{63.6} & \textbf{0.098}  \\
\bottomrule
\end{tabular}
\end{adjustbox}

\caption{\textbf{Moving Object Masking.} Ablation study on the IoU threshold used to determine if an object is moving. \textit{Linear} means that the IoU is decreased linearly from 0.7 with each training iteration.}
\label{mom_sup}
\end{table}

%%%%%%%%% BODY TEXT - ENTER YOUR RESPONSE BELOW

\begin{table}[h!]
\center	
\ra{0.9}
\begin{adjustbox}{max width=\textwidth}
\begin{tabular}{@{}l|c|c|c}
%\specialrule{.1em}{.05em}{.05em} 
\toprule
Model    & absRel $\downarrow $ & sqRel $\downarrow $ & RMS $\downarrow $ \\  \midrule %\specialrule{.1em}{.05em}{.05em} 
MTL Self-Supervised Depth & 0.106 & 0.841  & 5.270  \\  
+ Loss Balancing & 0.102 & 0.767 & 5.034 \\ \midrule
+ $\mathcal{L}_{PGS}$ & 0.101 & 0.781 & 5.010 \\ \midrule
+ $\mathcal{L}_{PED}$ &  0.101 & 0.778 & 5.024 \\ \midrule
+  $\mathcal{L}_{PGT}$ & 0.100 & 0.757 & 4.998 \\ \midrule
%+ $\mathcal{L}_{PGS}$ + $\mathcal{L}_{PED}$ & 0.101 &  0.789  &   4.997 \\
+  $\mathcal{L}_{PGS}$ + $\mathcal{L}_{PED}$ + $\mathcal{L}_{PGT}$ & 0.099 & 0.747  &   4.988 \\ 
+ Moving Objects Masking & 0.098 & 0.701 & 4.864 \\ 
\textbf{+ Extended dataset} & \textbf{0.082} &  \textbf{0.515}  &   \textbf{4.198} \\
\bottomrule
\end{tabular}
\end{adjustbox}
\caption{\textbf{Panoptic-guided Depth Evaluation.} Results on Cityscapes-DVPS. Ablation study for panoptic-guided depth losses and moving objects masking. }
\label{pan_depth2_sup}
\end{table}

\section*{Implementation Details} \label{impl_sup}

We adopt the ResNet-50 \cite{ResNet} backbone for the depth-aware video panoptic segmentation network. The network is pretrained on the Cityscapes dataset \cite{Cityscapes} for image panoptic segmentation. The pose estimation network follows \cite{monodepth2} with a ResNet18 backbone and a decoder that predicts the 6DOF camera pose, the translation vector and rotation matrix, as Euler angles. During inference we discard the pose estimation network. During training, we employ a minibatch of 4 images for 30k iterations, using the Adam optimizer with a base learning rate of $1e-3$ for decoders and heads and  $1e-4$ for the backbone and polynomial learning rate decay. We adopt image augmentation, such as random horizontal flip and random color augmentation: brightness, contrast, saturation and hue jitter. We employ image resolution $1025 \times 2049$ for Cityscapes-DVPS and $385 \times 1281$ for SemKITTI-DVPS. For depth evaluation, we center crop the Cityscapes-DVPS image to  $512 \times 1664$, in order to discard the sky and ego-vehicle regions, following \cite{monodepth2}.

\end{document}